\pdfoutput=1

\documentclass[11pt,hyphens]{article}

\usepackage[final]{acl}

\usepackage{latexsym}
\usepackage{times}

\usepackage{soul}
\usepackage{graphicx}

\usepackage{booktabs}
\usepackage{comment}

\usepackage{subcaption}
\usepackage[T1]{fontenc}

\usepackage[utf8]{inputenc}

\usepackage{microtype}

\title{Contextualizing Emerging Trends in Financial News Articles}

\author{Nhu Khoa Nguyen \\
  University of La Rochelle  \\
  F-17000, La Rochelle, France \\
  \texttt{\small{nhu.nguyen@univ-lr.fr}} \\\And
  Emanuela Boros \\
  University of La Rochelle\\
  F-17000, La Rochelle, France \\
  \texttt{\small{emanuela.boros@univ-lr.fr}} \\\And
  Gaël Lejeune \\
  Sorbonne University \\
  F-75006, Paris, France \\
  \texttt{\small{gael.lejeune@sorbonne-universite.fr}} \\\AND
  Antoine Doucet \\
  University of La Rochelle,  \\
  F-17000, La Rochelle, France \\
  \texttt{\small{antoine.doucet@univ-lr.fr}} \\\And
  Thierry Delahaut \\
  La Banque Postale Asset Management \\
  F-75004, Paris, France \\
  \texttt{\small{thierry.delahaut@labanquepostale-am.fr}} \\}

\begin{document}
\maketitle
\begin{abstract}
Identifying and exploring emerging trends in news is becoming more essential than ever with many changes occurring around the world due to the global health crises. However, most of the recent research has focused mainly on detecting trends in social media, thus, benefiting from social features (e.g. likes and retweets on Twitter) which helped the task as they can be used to measure the engagement and diffusion rate of content. Yet, formal text data, unlike short social media posts, comes with a longer, less restricted writing format, and thus, more challenging. In this paper, we focus our study on emerging trends detection in financial news articles about Microsoft, collected before and during the start of the COVID-19 pandemic (July 2019 to July 2020). We make the dataset accessible and we also propose a strong baseline (\emph{Contextual Leap2Trend}) for exploring the dynamics of similarities between pairs of keywords based on topic modeling and term frequency. Finally, we evaluate against a gold standard (Google Trends)  and present noteworthy real-world scenarios regarding the influence of the pandemic on Microsoft. 


\end{abstract}

\section{Introduction}
\label{sec:introduction}

Digital news, through many means of diffusion (online publishing platforms, social media, blogs, etc.) is considerably influential, as it not only shapes and forms public opinion but can also be a factor in the decision-making process of many industries that uses technology to improve activities and performance. Therefore, discovering hidden themes and trends residing in news data is essential to improve analyzing and managing development directions for many companies. The importance of identifying new trends before they emerge is further emphasized with the world-changing surrounding the health crisis triggered by the COVID-19 pandemic.

Emerging trend detection is the task of automatically extracting topics that are gaining attention and on the verge of being trending \cite{QuiDang2016}. Emerging topics usually indicate contents that are more popular in a short period, while growing in interest and utility over time. Moreover, topics that become a trend can either be short-lived or last for a long time depending on the nature of the event (e.g., traffic accidents, natural disasters, election campaigns, regulation enforcement, etc.).

Based on the platform of publication, the data used for detecting emerging trends can be classified into two classes: social media text and formal text. Corpora crawled from social media usually contain text that is short, concise and usually include social features (e.g. likes and retweets in Twitter) which benefits the task as they can be used to measure the engagement and diffusion rate of content. Because of this fact, data from social media has been extensively studied in various research \cite{Peng2018EmergingPT}.


However, formal text data, unlike the short sub-300 characters social media posts, comes with longer, less restricted writing format, yet does not include any social features (e.g., news articles, official documents, reports). Because of these differences, emerging trend detection on such data is rather under-researched. While there exist studies on financial data \cite{borsje2010semi,malik2011accurate} and news articles \cite{BangLiu2020}, most of them are based on  techniques such as latent Dirichlet allocation (LDA) topic modelling  \cite{Behpour2021,Bissoyi2020DiscoveringTT}, term frequency-inverse document frequency (TF-IDF) weighting technique \cite{Zhu2019,Enrico2020}, or different clustering types \cite{Cao2018,li2020topic} for the identification of either ``hot'' topics or emerging themes. 



Therefore, in this paper, our main contributions of our study are: {(1)} We build a dataset from the Bloomberg's Event Driven Feeds (EDF), containing news about Microsoft in the time interval from July 2019 to July 2020, which is the time period from six-month before and six-month after the COVID-19 outbreak. {(2)} We combine term TF-IDF and LDA for a more precise generation of keywords (bi-grams). {(3)} We utilize the latest contextual embeddings to represent the real temporality and variation of the semantics during different periods. {(4)} We make the dataset accessible along with the snapshot of Google Trends data used in our evaluation\footnote{The snapshot of Google Trends used in this paper can be found at: \url{https://github.com/nnkhoa/ms-edf-evaluation}. Please contact \url{thierry.delahaut@labanquepostale-am.fr} for the data. }.

The article is organized as follows. Section \ref{sec:related} discusses related work on emerging theme detection. Section \ref{sec:method} describes the data and explains our approach for detecting emerging trends in financial-based data and the experimental setup is established in Section \ref{sec:experiment} alongside with the evaluation metrics and the detailed analysis regarding the impact of the pandemic on Microsoft. 
Lastly, Section \ref{sec:conclusion} concludes the article with remarks and points to future work.


\section{Related Work}
\label{sec:related}

\paragraph{Trend Detection in Formal Datasets}
The general direction for trend detection in formal text data is to use statistical methods \cite{hughes-etal-2020-detecting,Daud2021FindingRS}, topic modeling \cite{Bolelli2009TopicAT,Behpour2021}, and clustering \cite{BangLiu2020,linger2020batch} 
Using financial business patents, the research by \citet{WonSangLee2017} aimed to identify emerging technology trends by applying latent Dirichlet allocation (LDA) with an exponentially weighted moving average of LDA probability, which affects whether a topic is ``hot'' or ``cold''. 
A refined version of TF-IDF, proposed by \citet{Zhu2019}, aims to discover ``hot'' topics according to ``hot'' terms based on time distribution information, user attention, and K-means clustering.  
Unlike previous work that tackled trend detection in official documents and newspapers, others focused on proposing new approaches using research and scientific papers, documents that generally contain citations and bibliographies that could be considered as additional features  
\cite{Nie2017,Xu2019,He2009DetectingTE,Israel2020}.  
However, exploiting bibliographies can have disadvantages in timeliness and content analysis, as discussed by \cite{Dridi2019}. The authors further proposed an approach, called Leap2Trend using \textit{temporal} word embedding that was generated by being trained on the data in an initial period of time, and then fine-tuned in the upcoming time frames. 
This approach also tracked the similarities between pairs of keywords over time, which yielded results suggesting the robustness and timeliness characteristics of the Leap2Trend. 

\paragraph{Detecting Trends in COVID-19 Pandemic}

Recent works have also been conducted within the period of the COVID-19 pandemic to study emerging trends within certain communities and gauge the impact of the outbreak through social media text \cite{Kassab2020OnNM}. \citet{Enrico2020} employed term-frequency analysis, calculate nutrition and energy metrics, while also using social features in order to extract hot terms and build a topic graph through co-occurrence analysis using Twitter data in Italy. 
Another research targeted peer-review papers regarding the COVID-19 virus and apply word embeddings and machine learning models to track novel insight surrounding the spreading of the virus \cite{Pal2021PredictingET}. 


\section{ Methodology}
\label{sec:method}

\begin{figure*}[ht]
  \centering
  \includegraphics[width=.97\linewidth]{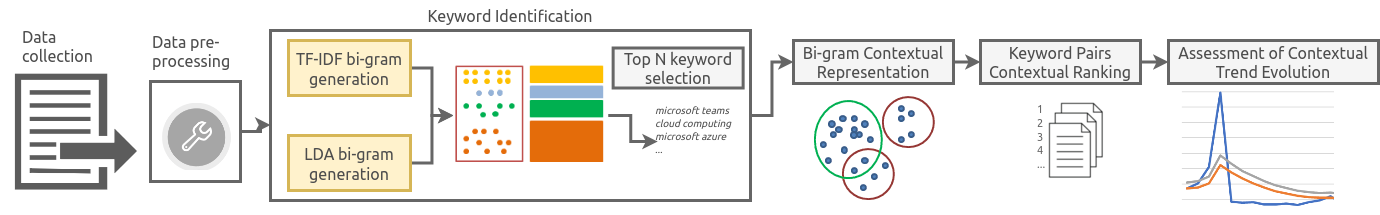}
  \caption{Summary of \emph{Contextual Leap2Trend}.}\label{figure:method}
\end{figure*}

For the current study, we built a dataset by extracting a portion of the Event-Driven Feeds (EDF) data, a proprietary dataset provided by Bloomberg\footnote{\url{https://www.bloomberg.com}}. Bloomberg L.P. provides financial software tools and enterprise applications such as analytics and equity trading platform, data services, and news to financial companies and organizations. The EDF data is massive and contains more than ten years of news collection that Bloomberg published in multiple languages from 2010 up until the present. 

\subsection{Data}

Considering the scope of the research, specific time frames and a company were chosen as follows: we studied the period from July 2019 to July 2020 - six months before the COVID-19 outbreak up until its peak, split into monthly collections of news. This is a very particular time span since the COVID-19 triggered a massive global health crisis and drastically changed people's lifestyle, which created new trends. 

For the specific company, we chose Microsoft which, at the time of writing this article, was known to play a major role in remote working trends during the pandemic by providing a stable and convenient platform for online workplace communication. Moreover, Microsoft is widely known as one of the leading companies in the field of cloud computing. Hence, it is interesting to investigate how trends in Microsoft shifted during the pandemic. It is important to note that while the choice of collecting news involving Microsoft was deliberate, any company could have been selected for our study given enough background knowledge about such company.

With these criteria, we extracted a total of 11,923 news articles about Microsoft within the said time span. 
The data has an enormous surge in the number of articles during the month of October 2019, which consisted of around 7,500 documents and accounted for more than 60\% of the total volume. In contrast, the number of documents during the beginning phase of the COVID-19 pandemic, in the span of 7 months from January 2020 to July 2020, is much lower in comparison. Averaging at about 300 articles per month, the period, the 7-month period only consisted just under 17\% of the total amount of articles in the dataset. With these figures, it is expected that October 2019 could be the month that Microsoft started a trend due to the huge spike in news articles volume, while the COVID-19 period may seem uneventful to the company.

\subsection{Data Pre-processing}
\label{subsec:pre-data}
From our analysis of the data, there are text patterns that appear relatively frequent in the corpus and mainly talk about Bloomberg's own publisher's detail, more exactly, the Bloomberg's standardized text as this provides no valuable knowledge to our task. Contrary, this standardized text could actually hinder the performance of a system as they could cause potential keywords to be more subtle, thus harder to be recognized. 
Moreover, since Bloomberg News is geared towards an audience that is interested in finance, articles usually contain an abundance of words that belong to the financial glossary (e.g. ``dividend'', ``bond'', ``equity''). The existence of this vocabulary could cause the same problem as with Bloomberg's standardized text, thus is it also removed from the text. Afterward, we perform a lemmatization to return to the canonical form of the token in the text, for reducing the size of the vocabulary to process in later steps. Lastly, any article that has less than ten tokens is considered uninformative and is removed\footnote{For this step, we manually checked different values of this threshold and the minimum quality of the remaining per-processed article.}.

\subsection{{Contextual Leap2Trend}}

Our approach adapts the Leap2Trend method proposed by \citet{Dridi2019} on detecting emerging themes in scientific papers, to financial-based documents in our case, and we refer to it as \emph{Contextual Leap2Trend}. 
The authors generated keywords representations were in different periods of time using static embeddings. 
Afterward, they assessed the similarity evolution of keyword pairs over time to depict which keywords are trending, thus forming topics based on the closeness between keywords. Leap2Trend also tackled the matter of lacking a gold standard to evaluate the result of trend detection by using Google Trends\footnote{\url{https://trends.google.com/trends/?geo=US}}. Google Trends collects search data of keywords and present them as interest rate over time, starting from 2004 to the present, which can be used to project emerging trends prediction results to gauge the performance of the system. 

Nonetheless, Leap2Trend has some disadvantages that needed to be addressed. First and foremost, Leap2Trend used a straightforward approach to extract the main keywords by inspecting titles of scientific papers for the most frequent bi-grams. The solution is justified by the fact that titles from scientific publications are written with the purpose of being self-explanatory and conveying the methods/problems clearly. The writing style leads to titles often containing a substantial amount of keywords. News articles, on the other hand, have condensed headlines that will only be expanded further in the main content of the documents, where most keywords reside. Thus, using raw frequency to extract keywords is inefficient due to noisy text overshadowing important phrases. Secondly, unlike in the scientific corpus that Leap2Trend used, where the context (which is mostly about the computer science field) is rather consistent, news collection, however, can contain numerous subjects ranging from technology, finance, economy to media. 


Thus, with \emph{Contextual Leap2Trend}, we propose to address the aforementioned disadvantages and to adapt them to our dataset an approach that is detailed in Figure \ref{figure:method} with the following steps: First, we pre-process our dataset to remove unwanted text in order to focus on the main content of the news article and divided the whole corpus into monthly sub-corpora (Section \ref{subsec:pre-data}). The next step identify potential keywords from the corpus by calculating the TF-IDF value as well as apply LDA to generate a list of top-rated bi-grams (Section \ref{subsec:kw}). Afterward, contextualized representations are generated for these trending keywords for each month (Section \ref{subsec:cntx}). We then rank the pair of keywords based on their representation similarity (Section \ref{subsec:ranking}). Lastly, we assess the contextual trend evolution in Section \ref{subsec:evo} by analyzing the change in ranking in each pair of keywords over each time span.

\subsection{Keyword Identification}
\label{subsec:kw}

The research on \textit{keyword identification} discovery originates from the topic detection and tracking (TDT) technology that was first studied by scholars in 1996 and its goal was making new detection and tracking within streams of broadcast news stories \cite{Zhu2019}. Emerging trends are usually signified by terms and phrases that are later considered as defining \textit{keywords} for such themes. Hence, correctly identifying potential keywords will lead to the right direction in discovering promising dormant trends. While keywords can be n-grams, in the scope of this research, only bi-grams were considered due to the fact that compared to uni-grams, they are less ambiguous, while appearing more frequently than other n-grams. Next, we present our methods of extracting potent keywords, using TF-IDF values to generate the list of highly important bi-grams, and utilizing LDA for getting bi-grams that can represent topics. 


\paragraph{TF-IDF Bi-gram Generation} 

TF-IDF captures how important a word is in the corpus by considering its frequency and penalizing it for appearing in too many entries in the corpus. Hence, words that are too common have considerably lower TF-IDF values than those that are less frequent. We exploited this method to extract important bi-grams from the corpus. Per month, TF-IDF values are generated for every bi-gram in the news collection and, afterward, we evaluated and produced lists of bi-grams that have either the highest average TF-IDF value in the collection or the highest single TF-IDF value across all documents. A high average TF-IDF value may indicate that a keyword associates closely with the company throughout the month, while a single high TF-IDF value signifies a sudden change in the context surrounding the company.

\paragraph{LDA Bi-gram Generation}

LDA derives from textual data the probabilities of words belonging to a predetermined number of topics. As such, the method excels at providing easily interpretable insights into what consists of a text corpus. Taking advantage of this fact, we used the results of applying LDA on the collection of texts in each month of our dataset to contextualize bi-grams by topics, which we obtained in the previous step. This is done by searching for topics that contain bi-grams in their list of words with the highest probability that represent topics. 
To find the optimal number of topics, we built numerous LDA models with different values of the number of topics and measured their topic coherence score \cite{Rder2015ExploringTS}. We chose the topic number that gives the highest coherence value. Coherence is a measure to evaluate to which degree the induced topics of an LDA model are correlated to one another, thus choosing the optimal number of topics that marks the end of the rapid growth of topic coherence usually offers meaningful and interpretable topics. 

\begin{figure}
    \centering
    \includegraphics[width=0.98\linewidth]{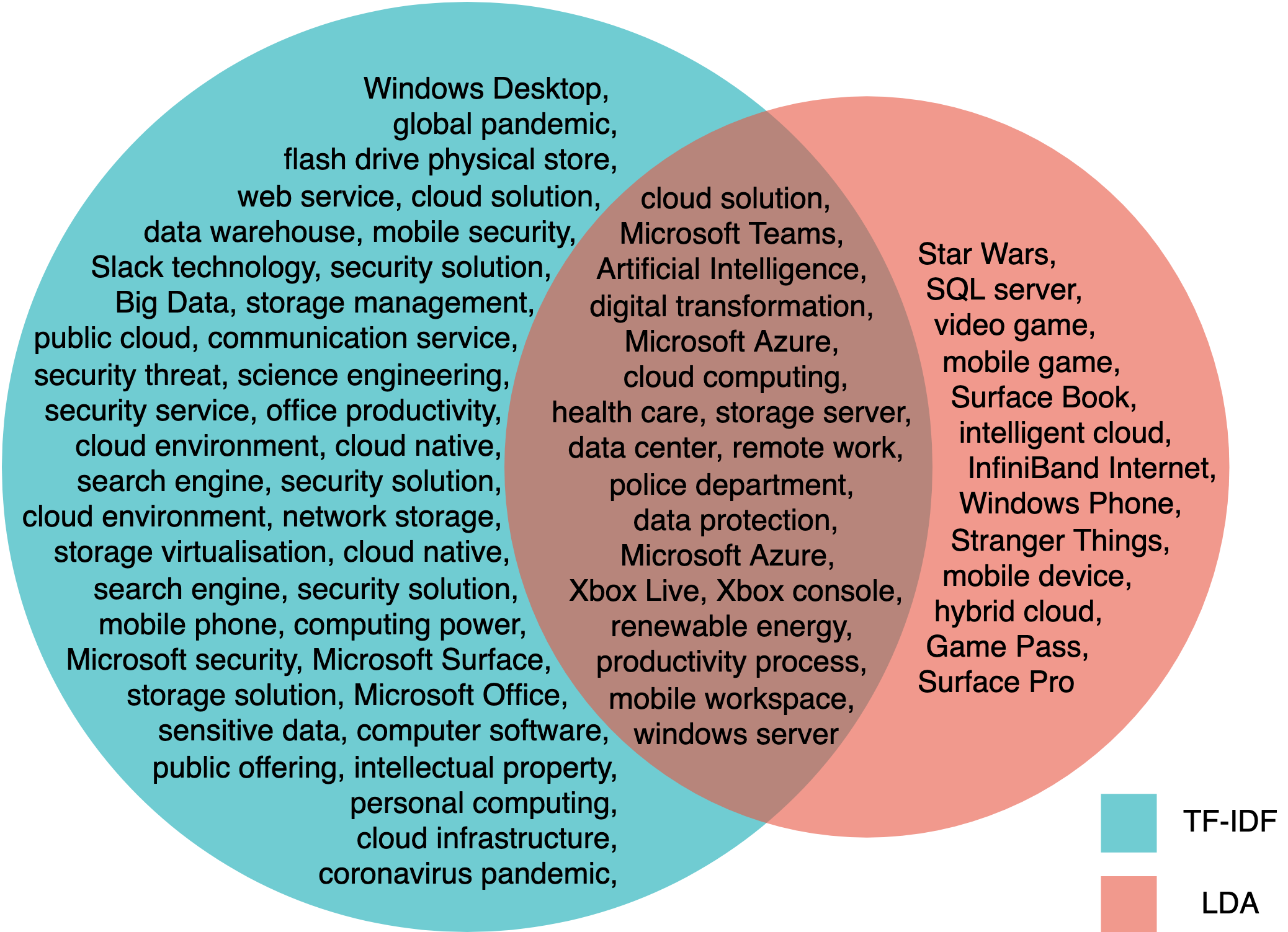}
    \caption{Keywords extracted by TF-IDF and LDA.}
    \label{fig:venn_tfidf_lda}
\end{figure}

\subsection{Bi-gram Contextual Representation Generation}
\label{subsec:cntx}

Unlike static word embeddings that capture the global representations of words in a vocabulary, contextual embeddings aim at representing each word or sub-word in the corpus depending on the words surrounding it. Therefore, each appearance of the word will have a unique vector assigned to it, which differentiates the same token but appears in context. For example, the token ``teams'' in ``Microsoft Teams'' and ``team management'' should be represented by distinct vectors, as their context and usage are entirely different. Thus, a contextual embedding is generally better than static word embedding, especially in a scenario where the corpus consists of news articles and not documents containing specialized knowledge. 

Because contextual embeddings usually derive vectors for one single token or a character n-gram separately, we employ the following strategy to generate bi-gram embedding to suit our need: for each occurrence of two tokens belonging to a bi-gram appears next to each other and in the right order, we generate FinBERT\footnote{\url{https://huggingface.co/ProsusAI/finbert}} embeddings for both tokens, and the final vector of the bi-gram is calculated by averaging them. 

\subsection{Keyword Pairs Contextual Ranking} 
\label{subsec:ranking}

With the contextual embeddings for the set of chosen keywords, we proceed to compute the similarity between each pair of keywords and rank them based on the value calculated. The idea is that when a number of terms appear frequently together, they usually share the same set of surrounding vocabulary, thus having similar context. 
For this, we computed the cosine similarity. Regarding how to establish the ranking, we first utilized the algorithm employed by {Leap2Trend} \cite{Dridi2019}, sorting the similarity of pairs of keywords in descending order, where the higher the similarity is, the lower the rank the pair of keywords has. 

\subsection{Assessment of Contextual Trend Evolution}
\label{subsec:evo}

Following the ranking calculation for each month, we attempted to assess the contextual evolution of each pair of keywords to identify potential emerging trends relating to the selected keywords. To achieve this, we analyze how much the rank increase/decrease between each month and set a threshold to decide whether changes in ranking signify emerging keywords that can lead to emerging trends. If the differences in ranking between the current month and the previous month of a pair of keywords are greater than the chosen threshold, we identify that this set of keywords will become the emerging terms. On the other hand, if the shift in ranking does not meet the threshold, the pair of keywords is regarded as having no potential in its emergence, either is falling off or is at standstill in terms of growth for being the next trend.

\section{Experimental Setup}
\label{sec:experiment}
We present the experiments on previously mentioned dataset from Bloomberg News about Microsoft from July 2019 to July 2020. This includes 
the description of the evaluation process, the results of our proposed system for emerging trend detection by comparing the shift in similarity ranking of pairs of keywords, and the elaboration of the story behind some keyword pairs that 
were in trend, and we deemed as interesting. 

\subsection{Evaluation Metrics}

To our knowledge at the current time of writing, there is no annotated dataset that is publicly available to experiment our method on. 
To produce a gold standard, the process involves examining Google Trends data, a platform for tracking terms/phrases popularity based on Google's search history, of the chosen keywords to identify where their emergence is. This was done by calculating the regression of interest rate evolution from a selected timestamp to \(N\) months forward, with a positive value indicating an increase in attention toward the keywords, thus signifying the possibility of the keywords belonging to emerging topics. Formula \ref{equ:regression} describes the regression of interest rate evolution in the next \(N\) months, denoted as \(m_{hits}\):

\begin{equation}
\label{equ:regression}
m_{hits}=\frac{\sum_{i=1}^{N}(x_i - \bar{x})(y_i - \bar{y})}{\sum_{i=1}^{N}(x_i - \bar{x})^{2}}
\end{equation}

where \(x_i\) and \(y_i\) represent the month number and the interest rate of that month, respectively. \(\bar{x}\) corresponds to the mean of the month number, while \(\bar{y}\) is the mean of interest rate.  

With the modality Google Trends treats a string in a search query, requesting data on just the literal phrase, we experienced difficulties in extracting interest rate data for combining a pair of keywords. Thus, we devised a solution by looking for data on each keyword separately and averaging the two results to get a final interest rate of the pair of keywords. The hypothesis behind this is the assumption that both keywords are part of the same theme/topic, thus, their evolution should have a similar tendency to rise/fall, albeit having different magnitude, making the combined signal stays relatively close to the two originals in terms of signal progression. 
Vice versa, phrases that do not fall into the same category cannot produce a good signal, which automatically makes them irrelevant to each other.

After obtaining the gold standard, we proceed to treat the task as a classification problem, where the system will classify whether changes in ranking context can lead to the same type of movement in the gold standard in the next $N$ months. Accordingly, the main evaluation metrics are precision, recall, and F1-measure. Not only do we consider the macro metrics, but we also take into account the aforementioned metrics on the true class detection specifically, since our focus is leaning toward correctly identifying emerging trends, which is signified by the true class. Additionally, we report the receiver operating characteristic (ROC) curves 
according to the selected time span onward with the purpose of assessing the detection capability on how many months forward can our system detect trends emergent effectively.

\subsection{Keyword Generation and Selection}



After the keyword generation step (bi-gram generation with TF-IDF and LDA), we noticed that while extracting bi-grams using raw TF-IDF value yielded many potential keywords (68 keywords), the amount of bi-gram that also existed in the LDA results was significantly lower (36 keywords). The intersection of the two lists resulted in a set of 22 keywords. The abundant amount of terms generated by TF-IDF and the high number of intersected keywords between two methods suggests that results from TF-IDF are less specific than that of LDA. From our observation, the semantics of keywords in the intersect region cover not only the general topics and trends such as health care due to COVID-19 pandemic, but also specific development direction of Microsoft. Figure \ref{fig:venn_tfidf_lda} details further on the list of extracted bi-grams and a total number of bi-grams yielded by each method.

\subsection{Results}
\label{sec:results}



We compare the performance of two systems: the original Leap2Trend that used monthly static Word2Vec embeddings and our \emph{Contextual Leap2Trend}. We set the threshold=$0$ for both system to imply that any positive change in context can signify potential emerging keywords. 

\begin{figure}[ht]
\centering
\begin{subfigure}{.45\textwidth}
 \centering
    \includegraphics[width=\linewidth]{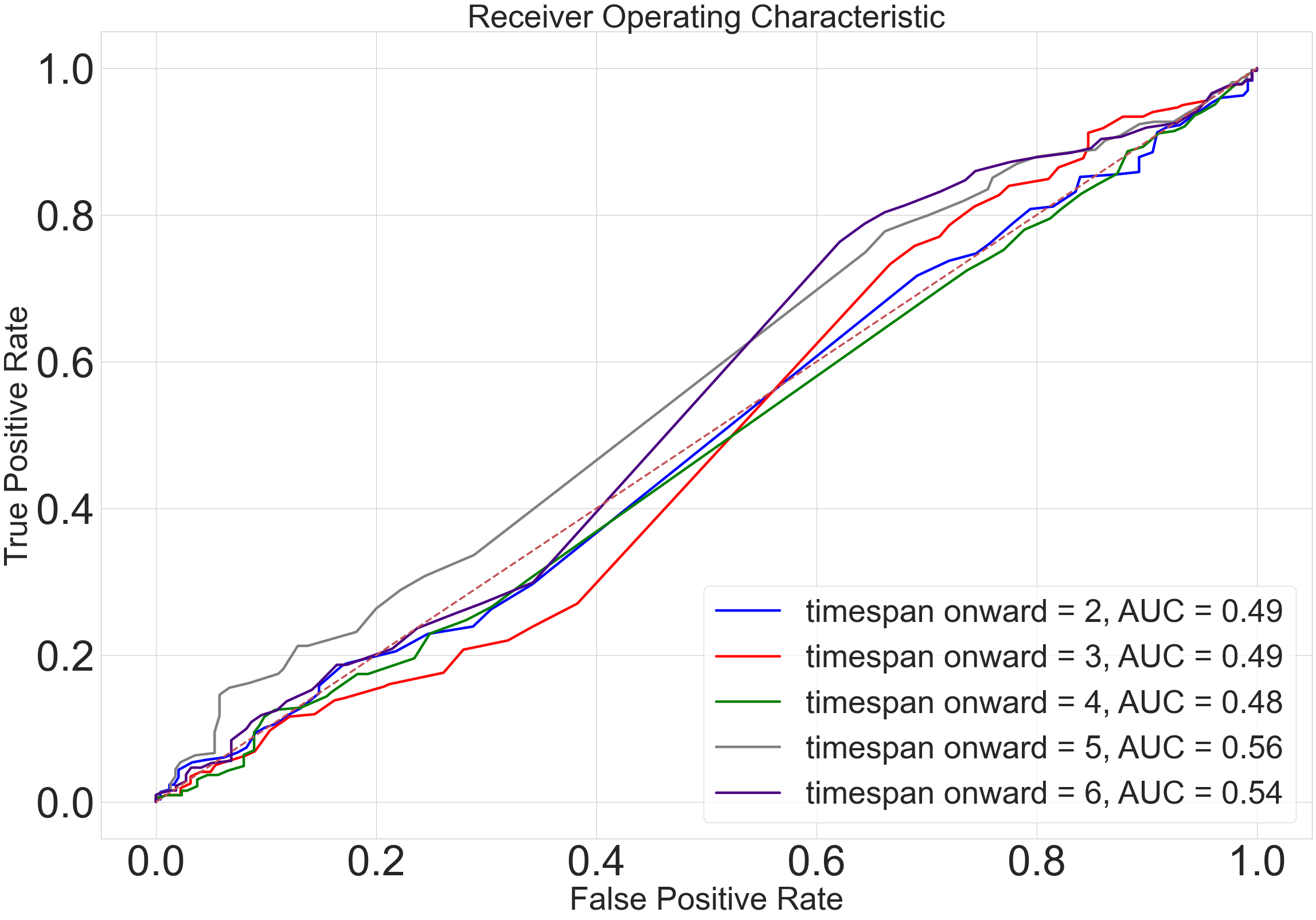}
 \subcaption{Original Leap2Trend\label{fig:roc_original}}
\end{subfigure}
\begin{subfigure}{.45\textwidth}
  \centering
    \includegraphics[width=\linewidth]{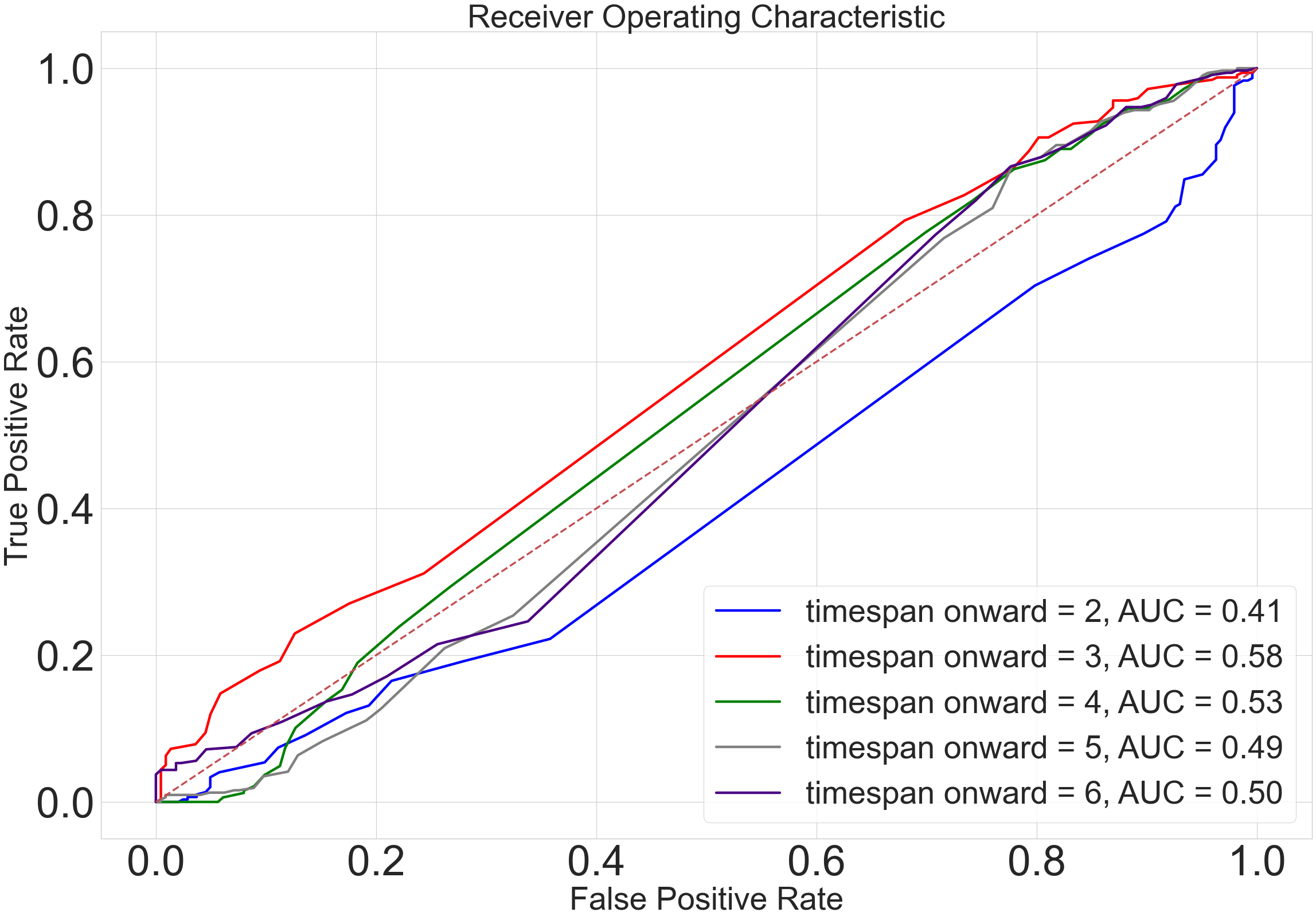}
  \subcaption{\emph{Contextual Leap2Trend} \label{fig:roc_contextual}}
\end{subfigure}
\caption{Compared ROC curves based on timespan adjustment (threshold = 0).}
\end{figure}

Figure \ref{fig:roc_contextual} demonstrates the predictive capability of what is the optimal number of months forward the \emph{Contextual Leap2Trend} respectively can perform the task efficiently. The two system outperformed one another in different timespan onward scenario, with the original Leap2Trend has better area under the curve (AUC) when assessing keywords trendiness potential within the span of $5$ and $6$ months($0.56$ and $0.54$ in AUC respectively). However, the \emph{Contextual Leap2Trend} system has the  AUC of $0.57$ when predicting 3-month forward which not only exceed the original system in the same category ($0.49$), but also is the best result overall. This observation suggests that by using contextual embeddings, our system surpasses the original Leap2Trend in terms of timeliness property which is crucial for the task of detecting emerging trend. In addition, the \emph{Contextual Leap2Trend} matches the original system in AUC ($0.54$) in the scenario of timespan forward=$6$.



\begin{figure}[ht]
\begin{subfigure}{\linewidth}
    \centering
    \includegraphics[width=.9\linewidth]{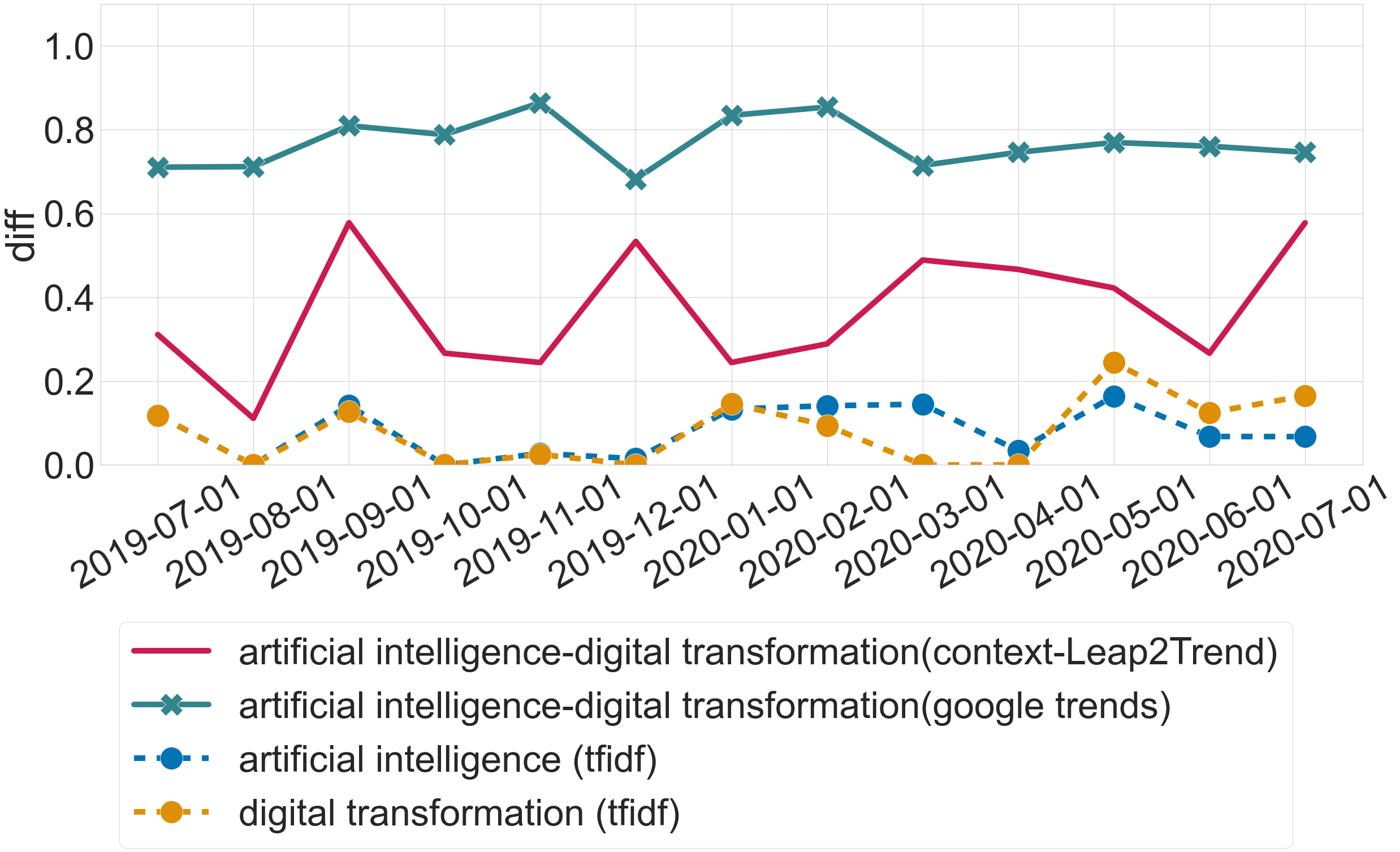}
    \caption{A.I./Digital Transformation}
    \label{fig:ai-dt}
\end{subfigure}
\begin{subfigure}{\linewidth}
    \centering
    \includegraphics[width=.9\linewidth]{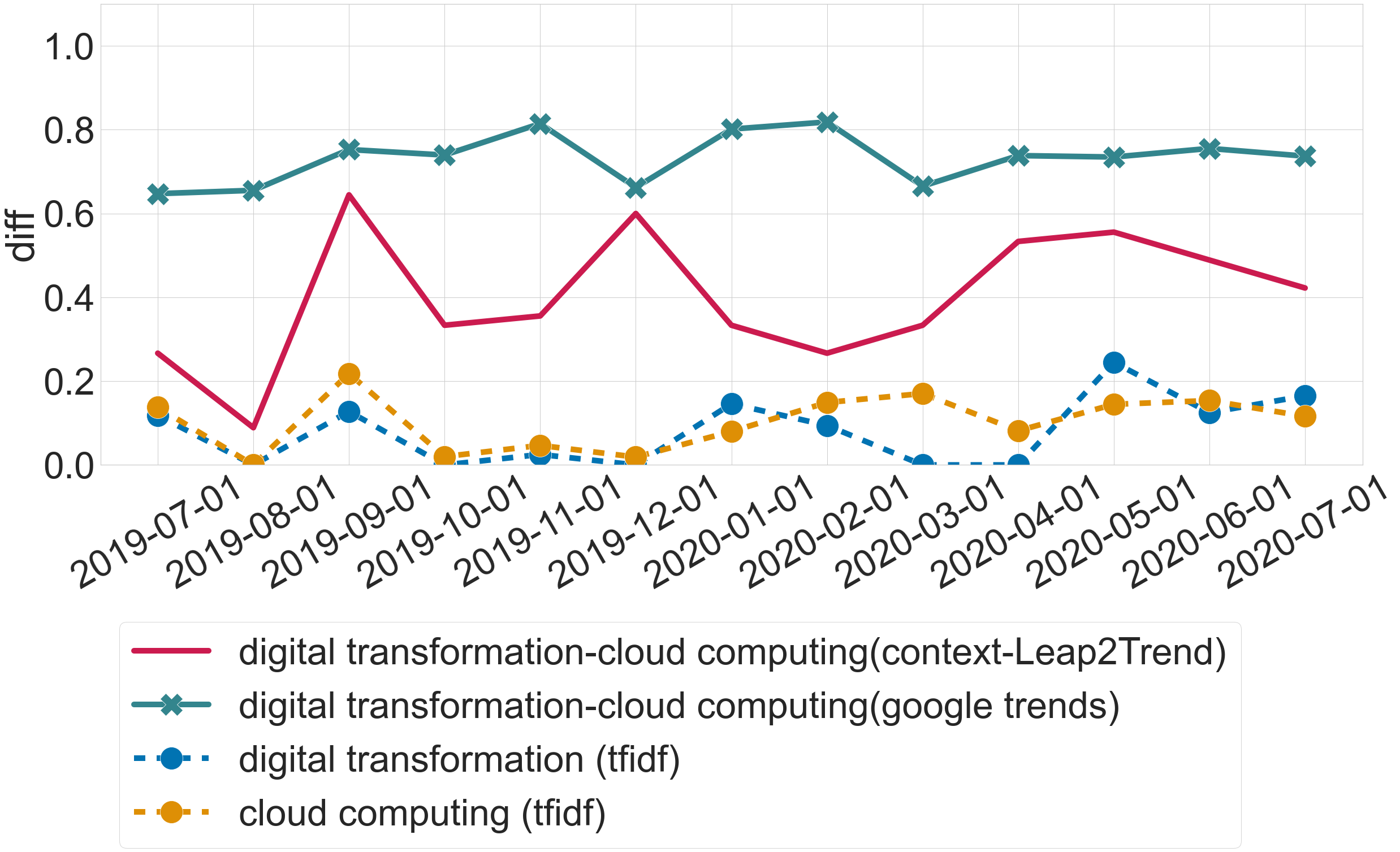}
    \caption{Digital Transformation/Cloud Computing}
    \label{fig:dt-cp}
\end{subfigure}
\begin{subfigure}{\linewidth}
    \centering
    \includegraphics[width=.9\linewidth]{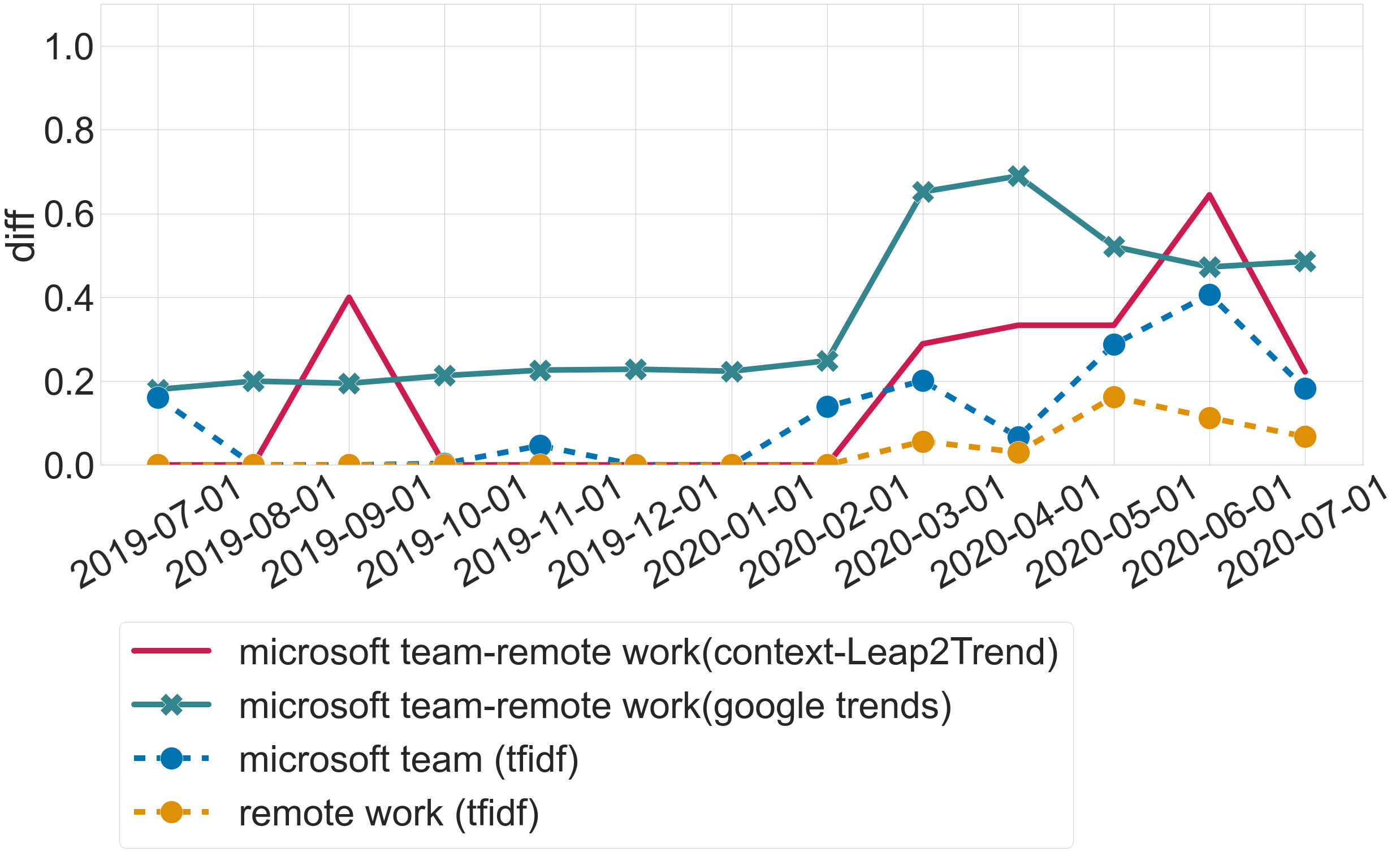}
    \caption{Microsoft Teams/Remote Work}
    \label{fig:mt-rw}
\end{subfigure}
\caption{Trends for the chosen pairs of keywords.}
\end{figure}


While the proposed system is more efficient at predicting trends three months in advance, with the current dataset, the task is considerably challenging. In order to gauge the performance of our system, we compared the results of a system with threshold of $0$ to a zero rule baseline where every possible bi-gram was considered as trendy (True class), thus True class metrics were not taken into account for this experiment. Table \ref{tab:zero-rule} shows that our system out-performed the zero rule baseline in all three metrics, but are only marginally better in terms of recall and F1 value. This observation further supports the difficulty of the task present using the current data.

Another observation is when comparing the change in ranking to Google Trends data, we notice that delays sometimes exists, usually of one month due to data split in the pre-processing phase, in how soon the ranking change with respect to Google Trends, meaning the timeliness aspect of detecting emerging trends might be affected. One possible explanation is that Bloomberg News could be behind in terms of timing compare to public response as Bloomberg mostly covers big events that were already happened, and does not cover innovations process that can lead to such event. 

In following sections, we discuss several example pairs of keywords that signified ongoing trends and emerging ones, mainly about what was happening surround the keywords while comparing the graph between Google Trends and contextual ranking evolution of \emph{Contextual Leap2Trend} system. 

\begin{table}[!ht]
  \caption{Proposed approach vs. zero rule baseline, timespan onward =3.}
  \label{tab:zero-rule}
  \setlength{\tabcolsep}{2.5pt}
  \begin{tabular}{p{1.5cm}cccc}
    \toprule
         \bf Method                      & \bf Thresh & \bf Precision & \bf Recall & \bf Macro F1\\
    \midrule
    Zero-rule baseline          &           & 0.3000 & 0.5000 & 0.3700\\
\midrule
Original Leap2Trend     & 0         & 0.5384 & 0.5330 & 0.5276 \\
\emph{Contextual Leap2Trend}          & 0         &  \bf 0.5600 &  \bf 0.5476 & \bf 0.5388 \\
    
  \bottomrule
\end{tabular}
\end{table}

\subsection{A. I. - Digital Transformation}

Compared to Google Trends data, for the \textit{artificial intelligence (A.I.) - digital transformation} pair of keywords, our proposed contextual ranking reflects their trends accordingly, as shown in Figure \ref{fig:ai-dt}. Digital transformation and artificial intelligence have been an ongoing conversation in technology and business in recent years as the movement seeks an effort to incorporate A.I. into digitizing business processes, customer experiences, etc. This is illustrated through the European Union's strategy to apply A.I. to digital transformation\footnote{\url{https://ec.europa.eu/commission/presscorner/detail/en/qanda\_20\_264}}. As for Microsoft, the company supports this trend with multiple projects, one of them being a major collaboration was mentioned in the EDF data\footnote{\url{https://www.bloomberg.com/press-releases/2020-03-26/c3-ai-microsoft-and-leading-universities-launch-c3-ai-digital-transformation-institute}}.

\subsection{Digital Transformation - Cloud Computing}

Within the context of the corpus, the \textit{digital transformation - cloud computing} pair of keywords describes how Microsoft's cloud computing service contribute to their involvement in advancing Digital Transformation with their business partners by providing Azure, Microsoft's leading cloud platform, to enhance the digital capability of their business partner\footnote{\url{https://www.bloomberg.com/press-releases/2020-04-07/blackrock-and-microsoft-form-strategic-partnership-to-host-aladdin-on-azure-as-blackrock-readies-aladdin-for-next-chapter-of}}. According to Figure \ref{fig:dt-cp}, the contextual ranking changes of the pair, in general, are in-line with Google Trends data, albeit displayed some level of differences. 

One thing to note is that it is the consensus, mentioned throughout the COVID-19 period in our corpus, regarding \textit{Digital Transformation} that because of the pandemic pushing society to stay distant and work remotely, the \textit{Digital Transformation} process will need to be developed faster to adapt to the current situation. In Figures \ref{fig:dt-cp} and \ref{fig:ai-dt}, it can be seen that this opinion were reflected through an increase in rankings starting March 2020. This period is also where the interest rate on Google Trends on this matter started to increase again after a dip.   

\subsection{Microsoft Teams - Remote Work}

While the magnitude displayed in Figure \ref{fig:mt-rw} was definitely lower than Google Trends's signal, Leap2Trend's signal visually still showed signs that the trends of the \textit{Microsoft Teams - remote work} pair of keywords have potential. Remote work, while being lesser known in 2019, has been a staple since the COVID-19 pandemic began. Zoom, a platform for online meetings, was booming at the start of the pandemic, yet fell in popularity due to security reasons. The slump of Zoom paved the way for the rise of Microsoft Teams as the reliable platform for workplace communication\footnote{\url{https://www.computerweekly.com/news/252485100/Microsoft-Teams-usage-growth-surpasses-Zoom}}. In our EDF dataset, the development of Microsoft Teams with new and better features also aid in its popularity\footnote{\url{https://www.bloomberg.com/news/articles/2020-03-19/microsoft-teams-boosts-work-at-home-effort}}.

\section{Conclusions}
\label{sec:conclusion}

This research addressed the drawback of existing methods in keyword extraction, bi-gram representation due to the differences in writing style between scientific papers and news articles. We, instead, introduced a combination of TF-IDF and LDA for generating potential keywords, and utilized contextual embeddings for the change in temporality. Our \emph{Contextual Leap2Trend} system showed considerable improvements compared to the original method in some scenarios in length of prediction. Moreover, we also presented several examples of emerging trends found in our data and the result also suggested that the approach has a good timeliness characteristic. 
In future work, we plan to introduce a better variety of data, such as news articles covering more companies and sector, to further experiment and improve our system. Moreover, increasing the time intervals could also uphold the consideration to assess trends longevity.

\bibliography{anthology,custom}




\end{document}